\documentclass[letterpaper, 10 pt, conference]{ieeeconf}  % Comment this line out if you need a4paper

\IEEEoverridecommandlockouts                              % This command is only needed if 
\usepackage{graphics} % for pdf, bitmapped graphics files
\usepackage{epsfig} % for postscript graphics files
\usepackage{mathptmx} % assumes new font selection scheme installed
\usepackage{times} % assumes new font selection scheme installed
\usepackage{amsmath} % assumes amsmath package installed
\usepackage{amssymb}  % assumes amsmath package installed
\usepackage{url}
\usepackage{booktabs}
\usepackage{multicol}
\usepackage{multirow}
\usepackage{hyperref}
\usepackage{cleveref}

\title{\LARGE \bf
Tracking-Aware Deformation Field Estimation \\for Non-rigid 3D Reconstruction in Robotic Surgeries
}

\author{Zeqing Wang$^{1,*}$, Han Fang$^{1,*}$, Yihong Xu$^{2}$ and Yutong Ban$^{1}$% <-this % stops a space
\thanks{This work has been supported by Shanghai Magnolia Funding Pujiang Program (No. 23PJ1404400) and SJTU-Xiaomi Scholar Funding.}% <-this % stops a space
\thanks{$^{1}$ Zeqing Wang, Han Fang and Yutong Ban are with UM-SJTU Joint Intuitute, Shanghai Jiao Tong University,
        Shanghai, China
        {\tt\small \{der-zing, han.fang, yban\}@sjtu.edu.cn}}%
\thanks{$^{2}$Yihong Xu is at Valeo.ai, Paris, France
        {\tt\small yihong.xu@@valeo.com}}%
\thanks{*indicates the authors contributed equally}% 
\thanks{Corresponding email: {\tt\small  yban@sjtu.edu.cn}}%
}
\usepackage{etoolbox}
\usepackage{color}
\usepackage{xcolor}
\newtoggle{final}
\toggletrue{final} % uncomment before submission

\newcommand{\methodfullname}{\textbf{T}racking-\textbf{A}ware \textbf{D}eformation \textbf{F}ield}
\newcommand{\methodname}{TADF}

\newcommand{\xyh}[1]{\iftoggle{final}{#1}{{\color{purple} #1}}}

\begin{document}

\maketitle
\thispagestyle{empty}
\pagestyle{empty}

%%%%%%%%%%%%%%%%%%%%%%%%%%%%%%%%%%%%%%%%%%%%%%%%%%%%%%%%%%%%%%%%%%%%%%%%%%%%%%%%
\begin{abstract}
% the problem and its importance
Minimally invasive procedures have been advanced rapidly by the robotic laparoscopic surgery. The latter greatly assists surgeons in sophisticated and precise operations with reduced invasiveness. Nevertheless, it is still safety critical to be aware of even the least tissue deformation during instrument-tissue interactions, especially in 3D space. 
% exsiting methods and problems.
To address this, recent works rely on NeRF to render 2D videos from different perspectives and eliminate occlusions. However, most of the methods fail to predict the accurate 3D shapes and associated deformation estimates robustly. Differently, we propose \methodfullname{} (\methodname{}), a novel framework which reconstructs the 3D mesh along with the 3D tissue deformation simultaneously. It first tracks the key points of soft tissue by a foundation vision model, providing an accurate 2D deformation field. Then, the 2D deformation field is smoothly incorporated with a neural implicit reconstruction network to obtain tissue deformation in the 3D space. Finally, we experimentally demonstrate that the proposed method provides more accurate deformation estimation compared with other 3D neural reconstruction methods in two public datasets. The code will be publicly available after the paper acceptance.

\end{abstract}
%%%%%%%%%%%%%%%%%%%%%%%%%%%%%%%%%%%%%%%%%%%%%%%%%%%%%%%%%%%%%%%%%%%%%%%%%%%%%%%%
\section{INTRODUCTION}

Robotic laparoscopic surgery has revolutionized minimally invasive procedures by enabling surgeons to perform intricate maneuvers with enhanced precision while minimizing invasiveness. Although commercial systems like the da Vinci Surgical System provide visualization tools, they \xyh{are costly} and remain fundamentally limited to passive observation of the surgical field. A critical challenge persists in intraoperative tissue deformation awareness: surgeons cannot physically contact with tissues through the robotic interface, potentially compromising safety during instrument-tissue interactions. 

Most procedures mainly face two limitations: (1) constrained by two-dimensional information within a confined perspective and (2) the occlusion of surgical instruments increases the complexity of tissue deformation analysis and may obscure critical information. These restrictions make surgeons rely on their own experience to judge the deformation as well as the applied force in certain scenarios, which comes with significant risks. For example, to learn and master video-assisted thoracoscopic surgery, surgeons need to perform 100-200 surgeries to achieve proficiency, and more than 200 surgeries to achieve lasting surgical stability \cite{PCM5025}. In such cases, if the deformation of the soft tissues can be estimated, it can largely facilitate the surgeons' recognition overload and provide a safety control.  

Recently, with the development of NeRF~\cite{mildenhall2020nerfrepresentingscenesneural}, which relies on deep learning to achieve 3D reconstruction,  many works \cite{endosurf,wang2024efficientendonerfreconstructionapplication} started to solve 3D representation problems for surgical scenarios by applying NeRF. 
Though those works do achieve significant progress on rendering high-quality videos from other perspective and are able to eliminate occlusion of surgical instruments, they %cannot 
\xyh{fail to} predict the dynamic 3D deformation accurately. More \xyh{precisely,}
%exactly, 
their model serves for rendering 2D videos, while for some 3D information, like deformation, their predicted values are inaccurate, according to our experiments. To tackle this issue, we propose \methodfullname{} (\methodname{}), \xyh{a novel method that }aims at predicting the 3D deformation by incorporating the key point tracking information into the neural deformation field. Our contribution can be summarized as follows:
 \begin{figure}[t]
    \centering
    \includegraphics[width=0.99\linewidth]{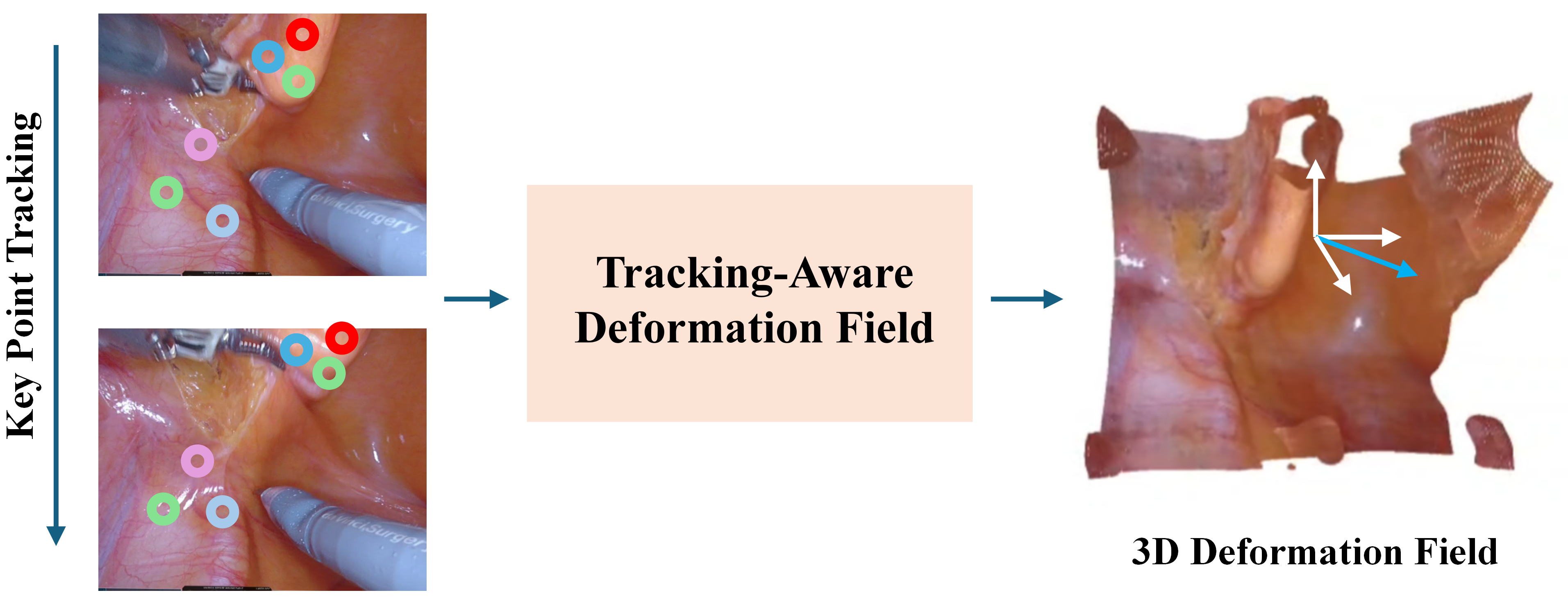}
    \caption{The proposed 3D Tracking-Aware Tissue Deformation Field (TADF) estimation framework. The colored circle represents each tracked key point.} 
    \label{fig:teaser}
\end{figure}
\vspace{-1ex}

\begin{itemize}
    \item We design a novel framework that estimates the three-dimensional deformation of the soft-tissue in robotic surgeries.
    \item The proposed framework contains a tracking module that tracks key points of tissue and incorporates the tracked displacement into the neural reconstruction network. This design guides the neural deformation network to predict the accurate deformation of the tissues.
    \item We design experiments to validate the performance of our model. Abundant experiments demonstrate that our model outperforms the vision-based 3D reconstruction model on both 3D reconstruction metrics and deformation metrics. 
\end{itemize}

\section{Related Works}\label{sec:related}

\subsection{3D Representation Methods for Surgical Scenarios}\label{sec:3d}

In our study, real-time dynamic reconstruction of surgical scenes forms the foundation for visual force field modeling. Due to the complex environment of surgical operations, characterized by limited space and deformable soft tissues, many 3D reconstruction methods cannot be applied effectively. The main challenges in this field include obtaining precise 3D models and achieving dynamic updates. SLAM methods, like~\cite{Song_2018}, through a point cloud-based data management approach and embedded deformation nodes to handle soft tissue deformation, three-dimensional reconstruction with an accuracy of 0.08–0.21 mm is achieved. Implicit methods, such as Neural Radiance Fields (NeRF)~\cite{mildenhall2020nerfrepresentingscenesneural}, represent the scene geometry and appearance through 5D neural radiance fields and support high-quality rendering. Other work~\cite{OrthogonalNeuralPlane} introduces Fast-Orthogonal-Plane (Forplane) based on NeRF which achieves efficient reconstruction and reduces reconstruction time from tens of hours to minutes. Stilz et al.~\cite{stilz2024flexjointposedynamic} propose a progressive optimization strategy that can handle long time scenarios, like 5000 frames.  

To address the specific needs of surgical scenes, improvements like EndoNeRF ~\cite{wang2024efficientendonerfreconstructionapplication} have been proposed, which incorporate specular reflection separation and instrument masking to reduce lighting interference from endoscopic video. However, NeRF’s training and rendering times are prohibitively long (ranging from hours to days for a single scene), making it unsuitable for real-time surgical applications. Recently, the reconstruction method based on 3D Gaussian splatting~\cite{kerbl20233dgaussiansplattingrealtime} has received great attention because of its high rendering speed and good structuring ability of the scene. Huang rt al.~\cite{huang2024endo4dgsendoscopicmonocularscene} raise lightweight MLPs to make the reconstruction more efficient and use the depth result from Depth-Anything~\cite{yang2024depthanythingunleashingpower} to generate geometry prior. On the other hand, in recent SurgicalGaussian~\cite{xie2024surgicalgaussiandeformable3dgaussians}, deformable 3D Gaussians~\cite{yang2023deformable3dgaussianshighfidelity} and deformation regularization constraints are used to ensure consistent motion of the local 3D Gaussian distributions, providing higher-quality details of the surgical process. Guo et al.~\cite{guo2024freesurgssfmfree3dgaussian} uses optical flow prior information to guide camera pose estimation and filtering outliers combined with consistency check, which achieves SfM-free 3DGS reconstruction in surgical scenarios.

% 3D Gaussian Splatting (3DGS)~\cite{kerbl20233dgaussiansplattingrealtime}, on the other hand, achieves efficient rendering ($\geq30 FPS$) by using 3D Gaussians, and dynamically adjusts anisotropic covariance to simulate tissue deformation (e.g., cutting or suturing). Other works, such as Deformable 3DGS ~\cite{yang2023deformable3dgaussianshighfidelity}, support real-time dynamic reconstruction by locally updating Gaussian parameters, providing high-quality, rapid reconstruction for monocular dynamic scenes. 

% Similarly, 3DGS has been applied to surgical scenes (e.g., ), where 

\subsection{Surgical Video Understanding}
The core objective of surgical video understanding is to integrate visual and geometric information to comprehend the surgical scene. Detection of surgical instruments and tissues is an essential task to achieve an accurate scene representation. Guru et al.~\cite{DetectRTools} uses a region proposal network to predict objectness and localization and achieve an average precision of ninety-one percent. Surgical phase recognition is an important field in surgical video understanding. Works ~\cite{LIU2025103366} introduce a Long Video Transformer to enhance phase recognition in long videos with similar frames. Liu et al.~\cite{Liu_2023_ICCV} introduces a fast Key information Transformer that predicts independently from the length of input frames and achieves high recognition accuracy. 

Semantic segmentation is another important task in understanding a surgical scene. One work~\cite{SHI2021102158} uses two-stage Semi-Supervised Learning strategy to segment surgical instruments with significantly improved segmentation accuracy. Holm et al.~\cite{DSG} utilize graph convolutional networks to achieve better understanding of interactions among tools in the whole surgical scenario. Other work~\cite{xu2024privacypreservingsyntheticcontinualsemantic} considers patient privacy and performs semantic segmentation by mixing open-source old instrumentation scenarios as a background while solving the issue of data scarcity. Yuan et al.~\cite{yuan2024segmentmodel2need} evaluate the performance of Segmentation Anything Model 2~\cite{ravi2024sam2segmentimages} in zero-shot surgery video and test its robustness. Some other methods use attention-based mechanism \cite{nwoye2022rendezvous} or graph-based neural networks \cite{yin2024hypergraph,ban2023concept} to infer the instrument-action-tissue triplets.

\subsection{Vision-Based Deformation Estimation}\label{sec:estimation}
% Vision-based Deformation is used to estimate the force feedback in some 
Intraoperative force sensing is crucial for providing force feedback in surgical robotics. 
Vision-based force estimation deduces the mechanical interaction forces through tissue deformation. 
Some studies~\cite{VisionBasedForceFeedback, Imageguidedsimulation} predict complex organ deformations by constructing biomechanical models and visually expressing the resulting forces, though these methods are computationally expensive and heavily dependent on accurate biomechanical parameters. A quantitative evaluation of the initial registration’s influence on the lesion location is essential. Some work~\cite{SequentialImageBasedAttentionNetwork} based on convolutional neural networks (CNNs) directly predict force signals from images circumventing the complexities of physical modeling.
However, those works neglect the occlusion of surgical tools, which in some cases results in false estimation.

While most modern surgical robots have enhanced visual feedback, traditional force feedback still faces limitations. In microsurgery, accurate force feedback is essential for guiding surgeons in fine motor control~\cite{HapticFeedbackandForceBasedTeleoperation}. One team~\cite{Haoyang2021ANV} uses specialized sensing devices to divide vascular intervention surgical robots into master and slave sides. The surgeon interacts with the control system through a master manipulator on the master side, while a slave wire feeder on the slave side collects force signals and sends them back to the control system. This setup enables force feedback during vascular intervention surgeries. Those methods usually require complex sensors, whereas visual feedback based methods only require cameras.

\section{METHOD}
\begin{figure*}[ht]
    \centering
    \includegraphics[width=0.8\linewidth]{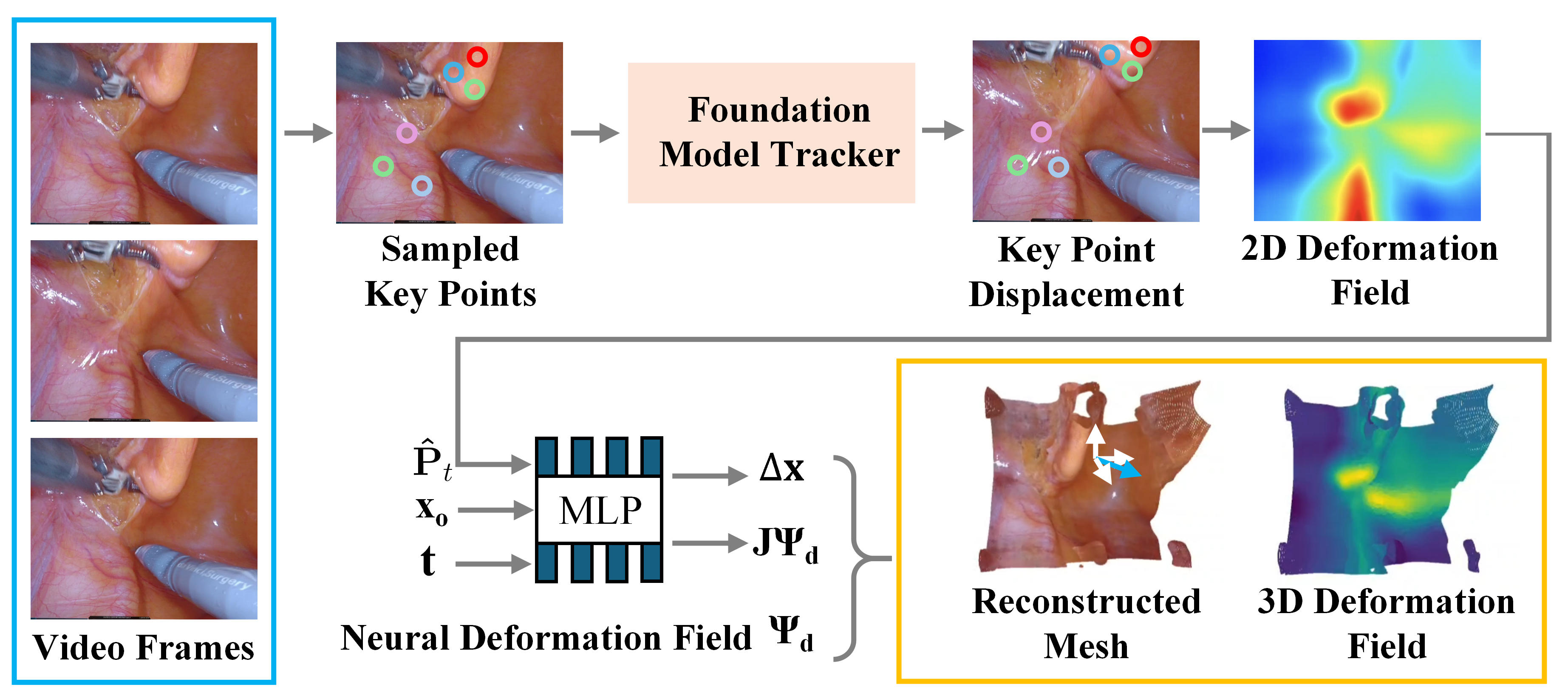}
    \caption{Overview of \methodname{}. For input video frames, we first sample key points and track those sampled points with the foundation CoTracker model. Taking the key point displacement as an additional input, our designed neural deformation field could predict an accurate deformation of tissues.}
    \label{fig:overview}
\end{figure*}
\subsection{Problem Setting}\label{sec:setting}
Given a video sequence of the deformation tissues, our goal is to estimate its displacement $\Delta x$ in three-dimensional space from its reference position $x_o$.
Following the notation of previous works~\cite{endosurf}, we receive a sequence of frame data $\{(\mathbf{I}_{i}, \mathbf{D}_{i}, \mathbf{M}_{i}, \mathbf{P}_{i}, t_{i})\}_{i=1}^{T}$ as input, where $T$ is the sequence length and $\mathbf{I}_{i}\in\mathbb{R}^{H\times W\times3}$ and $\mathbf{D}_i\in \mathbb{R}^{H\times W}$ are the $i$-th left RGB image and depth map with height $H$ and width $W$.
Foreground mask $\mathbf{M}_{i}\in\mathbb{R}^{H\times W}$ excludes unwanted pixels, such as surgical tools, blood, and smoke.

\subsection{\methodfullname{}}

% As mentioned in \Cref{sec:3d}, previous works \hf{cite} on 3D representation do achieve significant progress in reconstructing surgical environments.
% However, those works focus on the visual representation of the reconstruction, which neglects deformation, which is a key feature in surgical robotic area.
As shown in \Cref{fig:overview}, we propose \methodfullname{} (\methodname{}), which incorporates the sampling and tracking techniques into neural representation to achieve more accurate deformation estimation.
Meanwhile, we also introduce a way to visualize the estimated deformation for online visualization of the deformed tissue during surgery. The overall method contains several components, which is introduced as follows:
% In this section, we will thoroughly investigate the key technologies used in our approach. 
% As shown in \hf{fig}. We will elaborate on it through the following key steps:

% \begin{itemize}
% \item Removing the surgical instruments

% \item Applying the tracking model

% \item Processing tracking results and draw heatmaps

% \end{itemize}

\paragraph{Soft Tissue Key Point Sampling and Tracking}
% Describe how to sample and track key points (Zeqing)
Given an input video sequence, we first sample key points and track it by a foundation cotracker model~\cite{karaev2024cotrackerbettertrack}.
Our key point sampling strategy employs a grid-based approach to comprehensively capture soft tissue deformation patterns. We initialize a uniformly distributed grid $\mathcal{G} \in \mathbb{R}^{H_g \times W_g}$ across the surgical field, where $H_g$ and $W_g$ denote the grid dimensions in height and width respectively.

After sampling key points, we predict the 2D deformation of the soft tissues using the CoTracker~\cite{karaev2024cotrackerbettertrack} tracking model, a transformer-based tracking method that tracks multiple 2D key points independently and continuously. It is robust to tracking occluded points as well as points moving out of the camera's field of view. These features make CoTracker particularly suitable for complex surgical videos, providing accurate deformation tracking results.

\paragraph{Deformation Heatmap in Image Plane}
% how to interpolate the tracking results
To incorporate the sampled key point tracking results to the neural deformation field, we need to adapt the sparse grid-based motion tracks to full video resolution.
Hence, we employ a bilinear interpolation approach. 
Given initial predictions $\mathbf{P} \in \mathbb{R}^{T \times H_g \times W_g \times 2}$ from CoTracker (where $T$ is sequence length, $H_g \times W_g$ is the grid resolution), we interpolate to target resolution $H \times W$ through normalized grid sampling:

\begin{equation}
\mathbf{\hat{P}} = \text{Interpolate}(\mathbf{P}, H, W)
\end{equation}

The interpolation process consists of three key steps:
\begin{itemize}
    \item Grid Normalization: Create a normalized coordinate grid $\mathbf{G} \in [-1, 1]^{H \times W \times 2}$ for the target resolution.
    \item Bilinear Sampling: For each temporal frame $t$, sample from the prediction grid using bilinear interpolation.
    \item Temporal Consistency: Maintain temporal coherence by applying identical sampling grids across all frames.
\end{itemize}
% we first create a normalized coordinate grid $\mathbf{G} \in [-1, 1]^{H \times W \times 2}$ for the target resolution
% Grid Normalization: Create a normalized coordinate grid $\mathbf{G} \in [-1, 1]^{H \times W \times 2}$ for the target resolution.
% \begin{equation}
% G_{i,j} = \left( \frac{2j}{W-1} - 1, \frac{2i}{H-1} - 1 \right), \quad \forall i \in [0,H), j \in [0,W)
% \end{equation}

% Bilinear Sampling: For each temporal frame $t$, sample from the prediction grid using bilinear interpolation:
% \begin{equation}
% \mathbf{\hat{P}}_t[i,j] = \sum_{\alpha,\beta \in \{0,1\}} w_{\alpha\beta} \mathbf{P}_t[\lfloor x' \rfloor + \alpha, \lfloor y' \rfloor + \beta]
% \end{equation}
% where $(x', y')$ are the unnormalized grid coordinates:
% \begin{equation}
% x' = \frac{(G_{i,j}^x + 1)(W_g - 1)}{2}, \quad y' = \frac{(G_{i,j}^y + 1)(H_g - 1)}{2}
% \end{equation}
% with weights computed as:
% \begin{equation}
% w_{\alpha\beta} = (1 - |x' - \lfloor x' \rfloor - \alpha|)(1 - |y' - \lfloor y' \rfloor - \beta|)
% \end{equation}

% Temporal Consistency: Maintain temporal coherence by applying identical sampling grids across all frames:
% \begin{equation}
% \mathbf{\hat{P}} \in \mathbb{R}^{T \times H \times W \times 2} = \big[ \mathbf{\hat{P}}_t \big]_{t=1}^T
% \end{equation}

\begin{figure*}[h]
    \centering
    \includegraphics[width=0.99\linewidth]{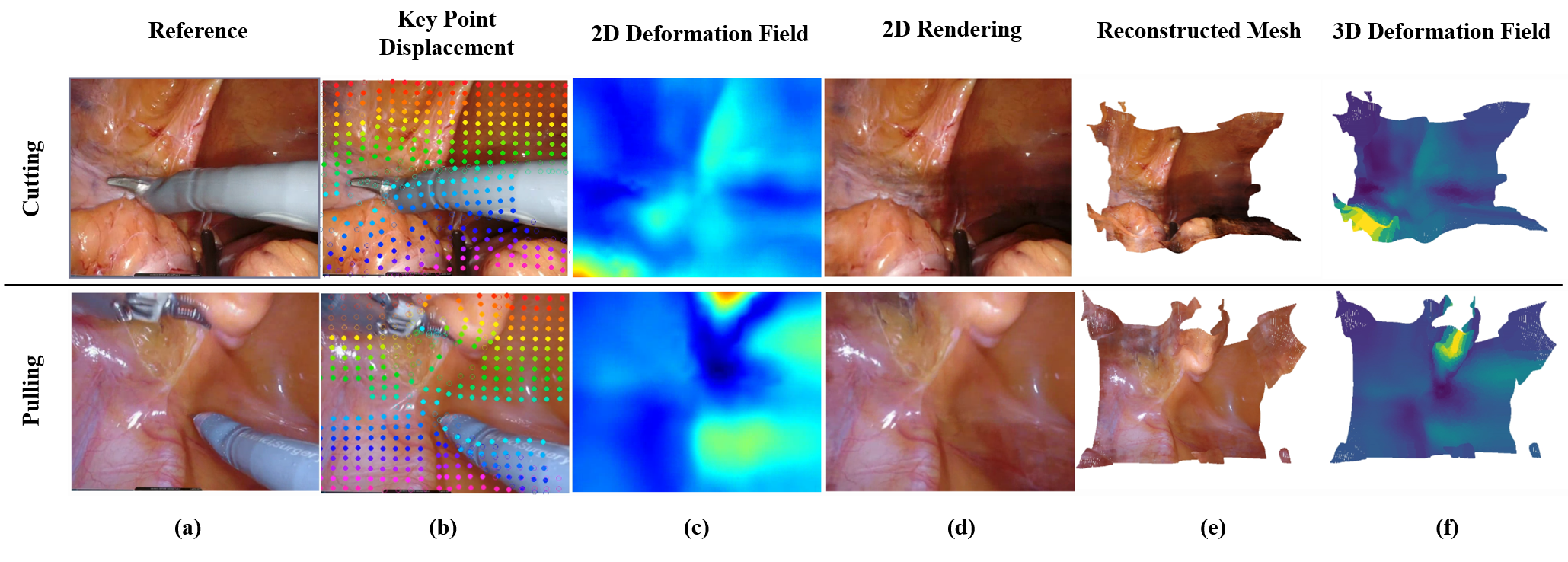}
    % \vspace{-1em}
    \caption{Intermediate Process Presentation for our Pipeline. Cutting means EndoNeRF-cutting dataset and Pulling means EndoNeRF-pulling dataset. Each subfigure represents a) reference - the input video frame, b) key point displacement - the sampled key points displacement tracked by foundation cotracker model, c) 2D deformation field - 2D visualization of deformation, d) 2D rendering - the rendering 2D videos after removing the surgical tools, e) reconstructed mesh - 3D reconstruction generated with the estimated deformation, and f) 3D deformation field - 3D visualization of deformation.}
    % \vspace{-2em}
    \label{fig:intermediate}
\end{figure*}

\paragraph{3D Deformation Estimation with Implicit Neural Representations}
% In most mainstream surgical videos, the presence of surgical tools significantly affects the accurate prediction of soft tissue deformation by tracking models. Therefore, how to remove surgical tools while maintaining the original deformation of soft tissues becomes a key challenge for research.

% To address this issue, we use the EndoSurf method, which is capable of reconstructing the 3D structure of a surgical scene. In the reconstructed 3D scene, after removing the surgical tools, we re-rendered the image sequence to obtain video frames without surgical tools but still retaining the original deformation of the soft tissue.
As mentioned in \Cref{sec:setting}, our goal is to estimate the deformation for a given stereo video of deformation issues.
However, in most mainstream surgical videos, the presence of surgical tools significantly affects the accurate prediction of soft tissue deformation by tracking models.
As mentioned in \Cref{sec:3d}, works like EndoSurf~\cite{endosurf} tackle this issue and are able to reconstruct videos with removal of surgical tools.
However, those works focus on the visual representation of the reconstruction, which neglects deformation, which is a key feature in surgical robotic area.
To overcome this problem, as shown in \Cref{fig:overview}, we use the interpolated tracking results $\mathbf{\hat{P}}_t$ of key points $\left\{p_i\right\}_{i=1}^N$ at the time step $t$, to guide the neural networks.
The neural deformation field could be represented as:
\begin{equation}
    \Psi_{d}(\mathbf{x}_{o},\mathbf{\hat{P}}_t,t)\mapsto \Delta\mathbf{x}.
\end{equation}
By transforming the raw view direction $\mathbf{v}_{o}$ using the Jacobian of the deformation field, which follows:
\begin{equation}
    \mathbf{J}_{\Psi_{d}}(\mathbf{x}_{o})=\dfrac{\partial \Psi_{d}}{\partial \mathbf{x}_{o}},
\end{equation}
we derive the canonical view direction $\mathbf{v}_{c}\in \mathbb{R}^{3}$ by
\begin{equation}
    \mathbf{v}_{c} = (\mathbf{I} + \mathbf{J}_{\Psi_{d}}(\mathbf{x}_{o})) \mathbf{v}_{o}.
\end{equation}

After calculating the deformation, we could get the spatial position $\mathbf{x}_c=\mathbf{x}_o + \Delta\mathbf{x}$. 
Following EndoSurf \cite{endosurf}, we construct the neural SDF field and neural radiance field to render color frames.
The optimization is executed by calculating two parts of the loss: rendering loss and geometry loss.
The rendering loss aims at minimize the difference between the rendered video and the ground truth video, while the geometry loss try to satisfy constraints on the neural SDF field.

\paragraph{Deformation Visualization}
Since we have obtained the 3D deformation and reconstructed mesh from previous steps, we want to demonstrate the deformation via a special visualization system:
\begin{equation}
\mathbb{D} = (\mathbf{M},\mathbf{x}_{o}, \Delta\mathbf{x}) \mapsto \mathbb{V} = (\mathbf{C})
\end{equation}
where $\mathbf{M}$ denotes the reconstructed mesh sequence, $\mathbf{x}_{o}$ and $\Delta\mathbf{x} \in \mathbb{R}^{T \times H \times W \times 3}$ are the spatial coordinates and the corresponding deformation, and $\mathbb{V}$ the visualization outputs containing color-encoded meshes $\mathbf{C}$.
The proposed visualization system integrates geometric processing with deformation analysis to enable an intuitive interpretation of soft tissue dynamics in surgical scenarios and it can output 3D deformation fields.

\begin{figure*}[ht]
    \centering
    \includegraphics[width=0.99\linewidth]{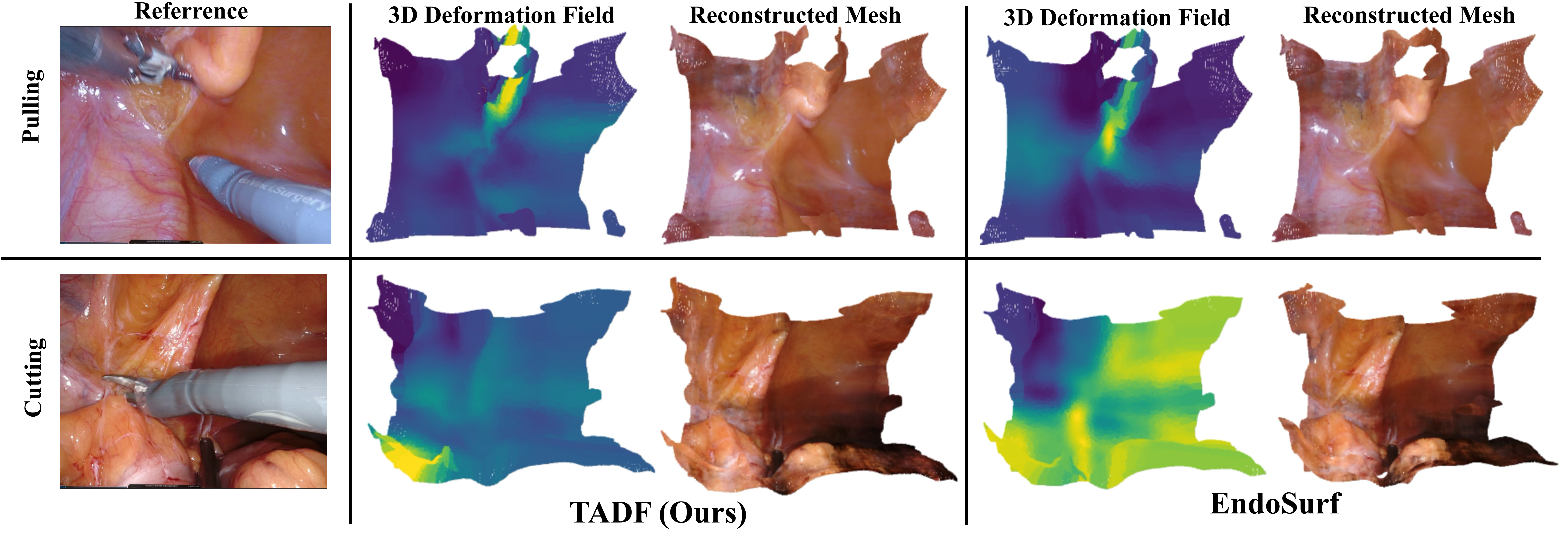}
    \caption{Visualization of Deformation and 3D reconstruction. We take one key frame in EndoNeRF-cutting and EndoNeRF-pulling datasets to visualize estimated deformation and 3D reconstruction.}
    % \vspace{-2em}
    \label{fig:visual}
\end{figure*}

\begin{table*}[ht]
    \centering
    \caption{Experimental Results of all the comparison methods on EndoNeRF and SCARED datasets.}
    \begin{tabular}{ccccccccccc}
    \toprule
     \textbf{Dataset}     &   \multicolumn{5}{c}{\textbf{EndoNeRF-cutting}} &  \multicolumn{5}{c}{\textbf{EndoNeRF-pulling}}  \\
     \multirow{2}{*}{Measurement}    &   \multicolumn{3}{c}{Vision Metric} & \multicolumn{2}{c}{Deformation Metric} &   \multicolumn{3}{c}{Vision Metric} & \multicolumn{2}{c}{Deformation Metric} \\
     & PSNR$\uparrow$ & SSIM$\uparrow$ & LPIPS$\downarrow$ & MaxSE$\downarrow$ & MSE$\downarrow$ & PSNR$\uparrow$ & SSIM$\uparrow$ & LPIPS$\downarrow$ & MaxSE$\downarrow$ & MSE$\downarrow$ \\
     \midrule
     \methodname{}&\textbf{35.411} & \textbf{0.954}&\textbf{0.104} & \textbf{0.273}& \textbf{0.334}& 36.226& \textbf{0.959}& \textbf{0.119}& \textbf{0.296} & \textbf{0.409}\\
     \methodname{} w/ noise& 35.088& 0.951& 0.112& 0.291& 0.358& 33.269 &  0.935 & 0.177 &0.368 & 0.460\\
     \cmidrule(lr){1-1}\cmidrule(lr){2-6}\cmidrule(lr){7-11}
     EndoSurf & 34.722& 0.945& 0.119 &0.296 &0.369 & \textbf{36.493} & 0.957& 0.121 &0.326 &0.427 \\
     EndoSurf w/ noise & 33.671& 0.931& 0.131& 0.341&0.402 & 31.492 & 0.898 &0.231 &0.391 &0.493 \\
     \midrule
     \textbf{Dataset}     &   \multicolumn{5}{c}{\textbf{SCARED-d1k1}} &  \multicolumn{5}{c}{\textbf{SCARED-d2k1}}  \\
     \multirow{2}{*}{Measurement}    &   \multicolumn{3}{c}{Vision Metric} & \multicolumn{2}{c}{Deformation Metric} &   \multicolumn{3}{c}{Vision Metric} & \multicolumn{2}{c}{Deformation Metric} \\
     & PSNR$\uparrow$ & SSIM$\uparrow$ & LPIPS$\downarrow$ & MaxSE$\downarrow$ & MSE$\downarrow$ & PSNR$\uparrow$ & SSIM$\uparrow$ & LPIPS$\downarrow$ & MaxSE$\downarrow$ & MSE$\downarrow$ \\
     \midrule
     \methodname{}& \textbf{24.292} &\textbf{0.738} & 0.375 & \textbf{0.301}& \textbf{0.234}& \textbf{32.241} & \textbf{0.963}& 0.069 &\textbf{0.202} &\textbf{0.175} \\
     \methodname{} w/ noise& 22.741 & 0.691 & 0.413 & 0.332&0.292 & 30.135& 0.951& 0.074 &0.253 &0.218 \\
     \cmidrule(lr){1-1}\cmidrule(lr){2-6}\cmidrule(lr){7-11}
     EndoSurf & 23.823 & 0.704 & \textbf{0.367} &0.345 &0.273 & 28.873 & 0.941 & \textbf{0.063} & 0.241 & 0.219 \\
     EndoSurf w/ noise & 20.172 & 0.655 & 0.431 &0.418 &0.354 &26.132 &0.912 & 0.085 & 0.295&0.273 \\
    \bottomrule
    \end{tabular}
    % \vspace{-2em}
    \label{tab:exp}
\end{table*}

The system first imports the reconstructed mesh sequences and the deformation results in parallel. Reshapes 2D deformation grids $\Delta\mathbf{x}$ into 1D arrays of 3D vectors while preserving spatial memory layout. Builds optimized spatial index structures for rapid nearest-neighbor search, utilizing parallel processing with multi-threaded query execution. Computes correspondence between mesh vertices $\mathbf{x}_{o}$ and deformation grid values $\Delta\mathbf{x}$ through minimum-distance matching. After the above procedure, the visualization of the deformation field can be obtained.

% We also apply a visual encoder to get heat map results. It first normalizes displacement magnitudes to [0,1] range using frame-specific extremal values and applies perceptually uniform Viridis colormap through linear RGB transformation. Then, it encodes color values as 8-bit integers for PLY format compatibility and we can get the 3D deformation fields based on reconstructed meshes.

\section{EXPERIMENTAL RESULTS}

\subsection{Experimental Setup}\label{sec:setup}
% Present the evaluation dataset, baseline methods, evaluation metrics, settings of experiments
\paragraph{Dataset}
We conduct experiments on two public endoscope datasets, EndoNeRF and SCARED.
EndoNeRF is composed by two videos, namely cutting and pulling.
Each of the video has an estimated depth map and a manually labeled tool mask.
For the SCARED dataset, we use its first two videos.
Each of the video contains ground-truth RGBD images of porcine cadaver abdominal anatomies.
Each of the datasets we use is preprocessed by normalizing the scene into a unit sphere and splitting the frames into 7:1 for training and testing.

\paragraph{Evaluation}
We have constructed two main evaluation metrics: vision metric and deformation metric.
Vision metric involves 3 different elements, which is Peak signal-to-noise ratio (PSNR), Structural Similarity (SSIM) and Learned Perceptual Image Patch Similarity (LPIPS).
For the deformation metric, we use the camera calibration matrix and depth image to estimate the ground truth pointcloud.
With the information of ground truth, we calculate the Mean Square Error (MSE) and Max Squared Error (MaxSE) to calculate the error between estimated deformation and ground truth.  

\paragraph{Baseline Methods}
We compare to the state-of-the-art method EndoSurf \cite{endosurf} as the baseline method to demonstrate the performance of our model.
Meanwhile, to demonstrate the robustness of our proposed \methodname{}, we also include ablation studies that randomly clear the input $\mathbf{x}_{o}$, where we denotes as "w/ noise" in \Cref{tab:exp}.

\subsection{Result Analysis}
% Compare our results with all the baseline methods with the given evaluation metrics
\Cref{tab:exp} demonstrates the comparison of our proposed \methodname{}  and EndoSurf across both EndoNeRF and SCARED datasets.

\paragraph{Overall Performance}
From the perspective of vision, our proposed \methodname{} achieves better results among all the metrics than EndoSurf on EndoNeRF-cutting dataset.
For the rest datasets, though our proposed \methodname{} does not achieve dominant results compared to EndoSurf, it always achieve better results on some metrics - SSIM and LPIPS on EndoNeRF-pulling, PSNR and SSIM on SCARED-d1k1 and SCARED-d2k1.
Taking the average performance, our proposed \methodname{} achieves an average PSNR of 32.043, SSIM of 0.904, LPIPS of 0.167, better than EndoSurf, which has an average PSNR of 30.978, SSIM of 0.887, LPIPS of 0.168.
This demonstrates that our proposed \methodname{} could even bring improvements on vision metrics, which could be attributed to the improvement accuracy of predicted deformation.

From the perspective of deformation, our proposed \methodname{} also outperforms EndoSurf, regardless of MSE and MaxSE.
In general, our proposed \methodname{} achieves an average max error of 0.268 and MSE of 0.288, better than EndoSurf, which has an average max error of 0.302 and MSE of 0.322.
It can be concluded that our proposed \methodname{} achieves our goal, which focuses on predicting the tissue deformation.

\paragraph{Robustness}
As mentioned in \Cref{sec:setup}, we also conduct experiments to demonstrate the robustness of our proposed \methodname{}.
\Cref{tab:exp} demonstrates that our proposed \methodname{} also has greater robustness than EndoSurf. 
Among all the datasets, under the noise perturbation, our proposed \methodname{} achieves better performance than EndoSurf on all metrics.
This illustrates that the additional input by CoTracker does not only improve the overall performance, but also enhance the robustness against noise.

\paragraph{Visualization Analysis}
\Cref{fig:intermediate} gives all the intermediates of our pipeline, which involves the reference, key point displacement, 2D deformation field, rendered 2D videos, reconstructed mesh and 3D deformation field.
As shown in \Cref{fig:visual}, we also provide the visualization comparison of the heatmap and 3D reconstruction generated by \methodname{} and EndoSurf.
For the deformation heatmap, compared to EndoSurf, our proposed \methodname{} could capture and emphasize the area that deforms most significantly.
For the frame in the cutting dataset, the EndoSurf estimated heatmap predicts severe deformation of a large area, while in the input video, only a small area of tissue deforms.
As for the pulling dataset, the deformation mainly happens on the raised tissue part, which is highlighted in our visualized deformation field.
However, in the deformation field generated by EndoSurf, the estimated deformation below this raised tissue turns out to be more significant, which violates the observation.
For the 3D reconstruction, the difference is not quite obvious on the static rendering frame.
However, in our generated video, our proposed \methodname{} provides smoother frames than what EndoSurf does.

\section{CONCLUSIONS}

\xyh{In this work, we propose a simple yet effective method  -- \methodname{} to estimate the tissue deformation of surgical stereo videos. This is achieved through the use of a point tracker that provides pixel-level displacements, which are incorporated into the neural deformation network through neural rendering. This novel approach achieves better accuracy than the state-of-the-art NeRF-based baseline on both EndoNeRF and SCARED public benchmarks, demonstrating its effectiveness and superior performance. This approach provides a feasible solution for safer surgical assistant systems with tissue deformation awareness. As for future work, we will address the efficiency through, for example, 3D gaussians~\cite{kerbl20233dgaussiansplattingrealtime} with real-time rendering. }

\addtolength{\textheight}{-12cm}   % This command serves to balance the column lengths

%%%%%%%%%%%%%%%%%%%%%%%%%%%%%%%%%%%%%%%%%%%%%%%%%%%%%%%%%%%%%%%%%%%%%%%%%%%%%%%%

%%%%%%%%%%%%%%%%%%%%%%%%%%%%%%%%%%%%%%%%%%%%%%%%%%%%%%%%%%%%%%%%%%%%%%%%%%%%%%%%


% \myTODO{optional}

\section*{ACKNOWLEDGMENT}

This work has been supported by Shanghai Magnolia Funding Pujiang Program (No. 23PJ1404400) and SJTU-Xiaomi Scholar Funding.

%%%%%%%%%%%%%%%%%%%%%%%%%%%%%%%%%%%%%%%%%%%%%%%%%%%%%%%%%%%%%%%%%%%%%%%%%%%%%%%%

\bibliographystyle{IEEEtran}
\bibliography{main}

\end{document}